\documentclass{article}

\pdfoutput=1
    \PassOptionsToPackage{numbers, compress}{natbib}

\usepackage[preprint]{neurips_2020}

\usepackage[utf8]{inputenc} 
\usepackage[T1]{fontenc}    
\usepackage{hyperref}       
\usepackage{url}            
\usepackage{booktabs}       
\usepackage{amsfonts}       
\usepackage{nicefrac}       
\usepackage{microtype}      
\usepackage{xcolor}
\usepackage{amsmath}
\usepackage{eufrak}
\usepackage{multirow}
\usepackage{wrapfig}
\usepackage{graphicx}

\usepackage{amsmath,amsfonts,bm}

\def\eqref#1{equation~\ref{#1}}

\def\1{\bm{1}}

\DeclareMathAlphabet{\mathsfit}{\encodingdefault}{\sfdefault}{m}{sl}
\SetMathAlphabet{\mathsfit}{bold}{\encodingdefault}{\sfdefault}{bx}{n}

\usepackage{algorithm}
\usepackage{algpseudocode}
\usepackage{subcaption}

\newtheorem{theorem}{Theorem}

\newcommand{\minus}{\scalebox{0.75}[1.0]{$-$}}

\title{Variance-reduced Language Pretraining via \\
a Mask Proposal Network}

\author{
  Liang Chen\thanks{Work is done during internship at Microsoft Research Asia.} \\
  Peking University \\
  \texttt{chanliang@pku.edu.cn} \\
}

\begin{document}

\maketitle

\begin{abstract}
Self-supervised learning, a.k.a., pretraining, is important in natural language processing. Most of the pretraining methods first randomly mask some positions in a sentence and then train a model to recover the tokens at the masked positions. In such a way, the model can be trained without human labeling, and the massive data can be used with billion parameters. Therefore, the optimization efficiency becomes critical.  In this paper, we tackle the problem from the view of gradient variance reduction. In particular, we first propose a principled gradient variance decomposition theorem, which shows that the variance of the stochastic gradient of the language pretraining can be naturally decomposed into two terms: the variance that arises from the sample of data in a batch, and the variance that arises from the sampling of the mask. The second term is the key difference between self-supervised learning and supervised learning, which makes the pretraining slower. In order to reduce the variance of the second part, we leverage the importance sampling strategy, which aims at sampling the masks according to a proposal distribution instead of the uniform distribution. It can be shown that if the proposal distribution is proportional to the gradient norm, the variance of the sampling is reduced. To improve efficiency, we introduced a MAsk Proposal Network (MAP-Net), which approximates the optimal mask proposal distribution and is trained end-to-end along with the model. According to the experimental result, our model converges much faster and achieves higher performance than the baseline BERT model. 
\end{abstract}

\section{Introduction}
In natural language processing, pre-trained contextual representations have been widely used to help downstream tasks that lack sufficient labeled training data. 
Previous works~\cite{radford2019language, yang2019xlnet, devlin2018bert,liu2019roberta} develop various self-supervised tasks to obtain pre-trained contextual representations. Taking the classic masked language modeling (MLM) task used by BERT \cite{devlin2018bert} as an example, it first randomly chooses a small number of positions in a sentence, mask the words on the position and then learns an encoder to restore them. As such tasks require no human supervision, the size of available training data could easily amount to the scale of billions of words. Pre-training over such large-scale data consumes exceptionally huge computational resources \cite{strubell2019energy}. 

In this paper, we tackle the training efficiency issue and develop a novel variance-reduced algorithm for better language pretraining. In particular, we observe that all previous works use uniformly sampled positions to mask when constructing their self-supervised tasks, and this is inevitably inefficient from the optimization perspective. For instance, in BERT training, we find that commonly used words and punctuation are easy to learn, i.e., those words (if being masked) can be correctly predicted by the model in just a few thousands of training steps. Meanwhile, some rare words and phrases are difficult to predict even at the end of the training. If we always uniformly sample the positions to mask, intuitively to say,  the variance of the stochastic gradient (with respect to the randomness of the masks) can be large since some positions gradually provide less informative signals while some do not. Usually, learning with a large-variance gradient estimator will be inefficient and ineffective. 

To formally characterize the variance of the stochastic gradient, we first introduce a principled gradient variance decomposition theorem. The theorem shows that the gradient variance can be naturally decomposed into two parts. One part concerns about the variance of sentences sampled in a batch, and the other part concerns about the variance of the masked positions. Our focus is on the variance reduction of the second part. Importance sampling is a standard way for variance reduction, which suggests that we can sample the masks using a proposal distribution instead of the uniform distribution. According to the theory, the variance is minimized if the probability of a mask sampled from the proposal distribution is proportional to the gradient norm. However, this brings the chicken-egg problem: We will never know the gradient norm unless we mask the positions, feed the masked sentence to the model and back-propagate the loss. As the number of possible masks is huge, feeding all the possibilities to the network to obtain their gradients is time expensive which significantly slows down the training process.

To address this challenge, we introduce a meta-learning approach by introducing a MAsk Proposal Network (MAP-Net) which takes the whole sentence as input and outputs a probability distribution over positions to sample masks. Both MAP-Net and the pretraining model are jointly optimized in an adversarial manner. Given a masked sentence sampled from the MAP-Net, the model is optimized to recover the masked sentence. At the same time, the MAP-Net receives signals from the performance of the model on this masked sentence, and improve itself. Instead of using reinforcement learning, we decouple the learning objective and make the training of the MAP-Net easier. We show that for language generation tasks, we can use the value of the loss instead of the value of the gradient norm. Therefore, the goal of the MAP-Net is to find ``tough'' masked positions with high losses to challenge the model, while the model attempts to fulfill the tasks generated by the MAP-Net. As we obtain the loss of many masked positions, the MAP-Net can be efficiently optimized from the pair-wise preference of different positions.  

To demonstrate the advantage of our proposed method, we conduct several experiments by using the MAP-Net to help the training of BERT, and evaluate them over GLUE natural language understanding benchmark \cite{DBLP:journals/corr/abs-1804-07461}. Experiment results first indicate that the masked words generated by MAP-NET are meaningful and informative during training. Furthermore, as the variance is sufficiently reduced, the BERT model learned with MAP-Net achieves better accuracy than the baselines on most of the tasks.

\begin{figure*}[t]
    \centering
    \includegraphics[width=5in, height=1.32in]{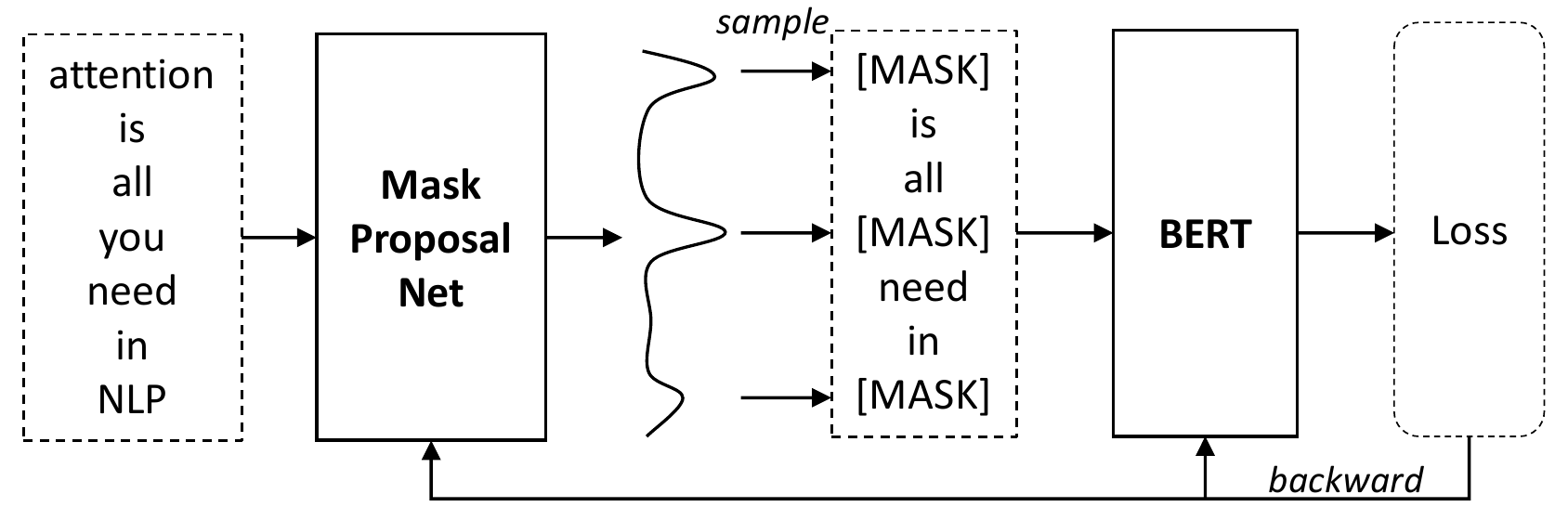}
    \caption{The learning framework.}
    \label{fig:overview}
\end{figure*}

\section{Related Work}
Pre-trained word vectors \cite{mikolov2013distributed,pennington2014glove} has been considered as a standard component in most NLP applications, especially for those tasks with a limited amount of labeled data~\cite{socher2011dynamic,tai2015improved,kalchbrenner2014convolutional}. However, one major weakness of such word vectors lies in its disregard on the rich syntactic and semantic structures of the word's surrounding context.

A few recent methods are proposed to learn pre-trained contextual representations by encoding the word's surrounding context, and the encoders are usually trained by a variety of self-supervised tasks using huge unsupervised data. For instance, \cite{peters2018deep, radford2018improving} train language models using stacked LSTMs \cite{hochreiter1997long} and Transformer layers \cite{vaswani2017attention}, and use the hidden states in the networks as the contextual representation. \cite{devlin2018bert,liu2019roberta} is well-known to use a masked language modeling task and achieves state-of-the-art performance on multiple natural language understanding tasks. Most recently, XLNet \cite{yang2019xlnet} and UniLM models \cite{dong2019unified} are proposed by designing permuted and bidirectional language modeling tasks.  

The exploding required computational cost, together with the resulting massive energy cost~\cite{strubell2019energy}, has become a great obstacle of applying deep neural network-based models in NLP. Unfortunately, to the best of our knowledge, there has been a limited number of works aiming at improving the training efficiency of such models.
\cite{you2019large} attempts to accelerate the training efficiency of the BERT model but it has to pay back with significantly-soaring computational resources.
\cite{gong2019efficient} observes that parameters in different layers have structural similarity, and the training time of BERT can be reduced using parameter sharing. A notable improvement is ELECTRA \cite{clark2019electra}, which proposes a small network to control the difficulty of the pre-training task.
And this is served as the starting point of our work.
\section{Method}
\subsection{Gradient Variance Decomposition of Language Pretraining}
Before elaborating the details of our proposed method, we first summarize the mathematical notations used in the following: we use $V$ to denote the word vocabulary and use $x = (x_1, ..., x_n)\sim P_X$ to denote a sentence of $n$ words, where $x_i\in V$, $i=1,2,\cdots,n$, and $P_X$ is the underlying distribution of the sentences; We use $x^{M}\sim\text{RandMask}(x)$ to denote a masked sentence of $x$ in which the RandMask operator first samples each position uniformly with some fixed probability, and then replace the sampled positions by a mask symbol \texttt{[MASK]}. 

We denote $\theta_{enc}$ as the parameter of the contextual encoder model. For most language pretraining tasks \cite{devlin2018bert,liu2019roberta}, using $x^{M}$ as input, the encoder learns to detect the difference between $x^{M}$ and $x$. Without loss of generality, we denote the loss function of as $\ell (\theta_{enc};x^M,x)$. For example, the masked language modeling loss can be formally written as
\begin{equation} 
\sum_{i: x^{M}_i=\texttt{[MASK]}} - \log P (x_i | x^{M},i;\theta_{enc})
,
\end{equation}
where  $P(v | x^{M},i;\theta_{enc})$ is the probability that the model predicts $v\in V$ as the missing word at the masked position $i$.

During training, we expect that the encoder performs well over the sentence distribution $P_X$ and all possible masks, and thus we introduce the \emph{expected loss} in which the expectation is taken over the randomness of the sentences and the masks.

\begin{equation} 
L(\theta_{enc}) = \operatorname{E}_{x\sim P_X}\operatorname{E}_{x^{M}\sim\text{RandMask}(x)}[\ell(\theta_{enc};x^M,x)]
,
\end{equation}

In practice, people seek to use stochastic gradient decent methods to optimize $L(\theta_{enc})$ and the stochastic nature of the methods induces randomness in the optimization. For simplicity, we use the single-batch setting for illustration of the stochastic gradient and its variance. All of the results can be easily extend to mini-batch setting. In single-batch setting, denote $g(\theta_{enc};x^M,x) = \frac{\partial \ell(\theta_{enc};x^M,x)}{\partial \theta_{enc}}$ as the stochastic gradient, where $x\sim P_X$ and $x^{M}\sim\text{RandMask}(x)$. We have the gradient variance decomposition theorem as below.

\begin{theorem}
Denote $\operatorname{Var}(g(\theta_{enc};x^M,x))$ as the variance of the stochastic gradient $g(\theta_{enc};x^M,x)$. It can be decomposed by the following formulation.
\begin{eqnarray}
\operatorname{Var}(g(\theta_{enc};x^M,x))&=& \operatorname{E}_{x\sim P_X}[\operatorname{Var}_{x^M\sim \text{RandMask}(x)}(g(\theta_{enc};x^M,x)|x)] \\
& +&\operatorname{Var}_{x\sim P_X}(\operatorname{E}_{x^M\sim \text{RandMask}(x)}[g(\theta_{enc};x^M,x)|x]),
\end{eqnarray}
\end{theorem}
The theorem above can be directly obtained by the law of total variance \cite{durrett2019probability}. From the theorem, we can see that the variance of the stochastic gradient is naturally decomposed into two terms. The first term characterizes the variance arising from mask sampling (we call it \emph{the mask variance}), and the second term characterizes the variance arising from sentence sampling (we call it \emph{the sentence variance}). In practice \cite{devlin2018bert}, a mini-batch of sentences, e.g., of size 512, is usually used for the gradient update and thus the sentence variance is not significant. However, for each sentence, only one mask is sampled, which can easily dominate the total variance. Therefore, in this work, we focus on the reduction of the mask variance.

\subsection{Importance Sampling via a Mask Proposal Distribution}

For Monte-Carlo methods, importance sampling is a standard way to reduce the variance when estimating the expectation of a random variable. Assume we have a data distribution $P_1$ and $f(\cdot)$ is the function of interest. To estimate the value $\operatorname{E}_{z \sim P_1}[f(z)]$, we can use a proposal distribution $P_2$ to sample $z$ independently and use the empirical average of sampled $\frac{p_1(z)}{p_2(z)}f(z)$ for the estimation, where $p_1, p_2$ are the probability density functions of $P_1, P_2$ respectively. If the distribution $P_2$ can be chosen properly, we will have a low-variance unbiased estimation of $\operatorname{E}_{z \sim P_1}[f(z)]$ according to the following theorem.

\begin{theorem}\cite{alain2015variance}
Denote $z_1,\cdots,z_T$ are i.i.d sampled from the proposal distribution $P_2$. Then $\frac{1}{T}\sum_t\frac{p_1(z_t)}{p_2(z_t)}f(z_t)$ is an unbiased estimator of $\operatorname{E}_{z \sim P_1}[f(z)]$. $\operatorname{Var}_{z \sim P_2}[\frac{p_1(z)}{p_2(z)}f(z)]$ is minimized if probability density function $p_2(z)\propto ~ \parallel f(z)\parallel_2$.
\end{theorem} 

Mapping back to our language pretraining scenario, the above theorem suggests that instead of using the $\text{RandMask}(x)$ operation that samples mask positions uniformly, we can design a non-uniform proposal distribution $\text{PropMask}(x)$  to sample $x^M$. For any sentence $x$, denote $p_{\text{prop}}(x^M|x)$ as the probability of $x^M$ with $\text{PropMask}(x)$, and $p_{\text{rand}}(x^M|x)$ as the probability of $x^M$ with $\text{RandMask}(x)$. Then using important sampling with $\text{PropMask}(x)$ operation, the loss we optimize can be rewritten as
\begin{equation} 
L(\theta_{enc}) = \operatorname{E}_{x\sim P_X}\operatorname{E}_{x^{M}\sim\text{PropMask}(x)}[\frac{p_{\text{rand}}(x^M|x)}{p_{\text{prop}}(x^M|x)} l(\theta_{enc};x^M,x)]
,
\end{equation}

and the variance of the stochastic gradient becomes
\begin{eqnarray}
\operatorname{Var}(g(\theta_{enc};x^M,x))&=& \operatorname{E}_{_{x\sim P_X}}[\operatorname{Var}_{x^M\sim \text{PropMask}(x)}(\frac{p_{\text{rand}}(x^M|x)}{p_{\text{prop}}(x^M|x)}g(\theta_{enc};x^M,x)|x)] + \nonumber\\
&&\operatorname{Var}_{x\sim P_X}(\operatorname{E}_{x^M\sim \text{PropMask}(x)}[\frac{p_{\text{rand}}(x^M|x)}{p_{\text{prop}}(x^M|x)}g(\theta_{enc};x^M,x)|x])
\end{eqnarray}

It can be easily seen that using $\text{PropMask}(x)$ does not change the sentence variance due to the fact that  $\frac{p_{\text{rand}}(x^M|x)}{p_{\text{prop}}(x^M|x)}g(\theta_{enc};x^M,x)$ is unbiased. As a direct adaptation of Theorem 2, we have for any possible $x^M$, if $p_{\text{prop}}(x^M|x)\propto \parallel g(\theta_{enc};x^M,x) \parallel_2$, the mask variacne is minimized.

Please note that it is costly to find the optimal $\text{PropMask}(x)$ during training. We can see that the optimal $\text{PropMask}(x)$ adaptively changes as $\theta_{enc}$ updates. More importantly, at each iteration, we require the value of $\parallel g(\theta_{enc};x^M,x) \parallel_2$ for each possible $x^M$ to determine the optimal $\text{PropMask}(x)$. However, for each sentence $x$, feeding all $x^M$ to the network to obtain the gradient norm is inefficient and will significantly slow down the training process.  

\subsection{Adversarial Training with a Mask Proposal Network}
In this subsection, we introduce our proposed method to deal with the challenge described above. We first deal with the computation of $\parallel g(\theta_{enc};x^M,x) \parallel_2$ and show that this value is positively co-related to the loss. For the BERT model, as shown in Eqn. (1), the loss for any masked position $i$ is defined as $- \log P (x_i | x^{M},i;\theta_{enc})$ and the gradient of $\theta_{enc}$ can be computed by the chain rule.
\begin{eqnarray}
\frac{\partial - \log P (x_i | x^{M},i;\theta_{enc})}{\partial \theta_{enc}}&=& -\frac{\partial P (x_i | x^{M},i;\theta_{enc})}{\partial \theta_{enc}}\times \frac{1}{P (x_i | x^{M},i;\theta_{enc})}\\
&=&-\frac{\partial P (x_i | x^{M},i;\theta_{enc})}{\partial \theta_{enc}}\times e^{-\log P (x_i | x^{M},i;\theta_{enc})}\\
&\approx&-\frac{\partial P (x_i | x^{M},i;\theta_{enc})}{\partial \theta_{enc}}\times (1 - \log P (x_i | x^{M},i;\theta_{enc}))
\end{eqnarray}
We can see that for such a likelihood loss function, the gradient is closely related to the scale of the loss, which suggests us using the loss instead of the gradient norm when seeking the optimal mask proposal distribution. The benefit of this approximation is two- folds: First, we don't need any additional computation to back-propagate the loss to the parameters to obtain the gradient norm, which is more efficient. Second, this makes the sentence-level task into position-level tasks, which is more tractable. The loss of a masked sentence is the average over the loss on the masked positions. We can decompose the task of finding a better $x^M$ for the sentence into determining which word is more preferred to be masked than another. As will be shown in the experiments, we find such an approximation is effective. 

Given this approximation, we are seeking a distribution that $p_{\text{prop}}(x^M,x)\propto \parallel l(\theta_{enc};x^M,x) \parallel$. Intuitively, we are looking for a mask distribution that can automatically sample some ``harder'' mask that induces a large loss for the encoder. This motivates us to learn a neural network that plays the role of learning to mask and jointly optimize it with the encoder in an adversarial manner. We call the neural work \emph{the MAsk Proposal Net} (MAP-Net).

To be concrete, denote $\theta_{\text{MAP}}$ as the parameter of the MAP-Net, which takes a sentence $x$ as input and outputs a multinomial distribution over the positions. For sentence $x$, we sample $K$ positions using the MAP-Net without replacements,
\begin{equation*}
     \text{pos}_1,...,\text{pos}_K\sim \text{MAP-Net}(x; \theta_{\text{MAP}}),
\end{equation*}
and obtain $x^M$ by masking the sampled position $\text{pos}_1,...,\text{pos}_K$. We simply approximate the likelihood ratio as $r(x^M) = \frac{p_{\text{rand}}(x^M|x)}{p_{\text{prop}}(x^M|x)} \approx \frac{(1/n)^K}{\Pi^K_{k=1} p_{\text{prop}}(pos_k)}$, which will be served as the weights on the loss. For sake of algorithmic stability, we clip the ratio $r_{clip}(x^M) =  \text{CLIP}(r(x^M), 1-\epsilon,  1+\epsilon)$ where $\epsilon>0$ is predefined. Then practically the loss on a sampled sentence $x$ is $r_{clip}(x^M) l(\theta_{enc};x^M,x)$, and $\theta_{enc}$ is updated to minimize this loss function over a mini-batch of sentences.

When feeding the masked sentence $x^M$ to $\theta_{enc}$, we also obtain some signals that can be used to update $\theta_{\text{MAP}}$. For each position $\text{pos}_k$, we have its loss $l(\theta_{enc};x^M,x,\text{pos}_k) =- \log P (x_{\text{pos}_k} | x^{M},\text{pos}_k;\theta_{enc})$. If $l(\theta_{enc};x^M,x,\text{pos}_k)$ is lower than the loss on other positions, the task of restoring $x_{\text{pos}_k}$ can be considered to be easier, and we should lower down the probability $p_{\text{prop}}(pos_k)$ given by MAP-Net. If $l(\theta_{enc};x^M,x,\text{pos}_k)$ is very high, the prediction task task on position $\text{pos}_k$ can be considered to be difficult. Therefore, we can update $\theta_{\text{MAP}}$ to increase $p_{\text{prop}}(pos_k)$. Based on such an intuition, we minimize the following objective function.

\begin{equation}
 L(\theta_{MAP}) = \sum^K_{k=1}-\log p_{\text{prop}}(\text{pos}_k))\cdot [l(\theta_{enc};x^M,x,\text{pos}_k)  - \text{baseline}], 
\end{equation}
where baseline is the average loss $\sum^K_{k=1}  l(\theta_{enc};x^M,x,\text{pos}_k) / K)$. Similar to Generative Adversarial Nets, $\theta_{enc}$ and $\theta_{\text{MAP}}$ are optimizing using Eqn (5) and (10) iteratively. The general training process can be found in Algorithm 1.

\begin{algorithm}
\caption{Proposed Algorithm}\label{alg}
\begin{algorithmic}[1]
\State \textbf{Input}: Sentence corpus $S$,  $\theta_{enc}$, $\theta_{MAP}$.
\Repeat
\State Sample a minibatch $\hat{S}$ from $S$.
\State Generate the masked positions for each sentence in $\hat{S}$ using $\theta_{MAP}$.
\State Update $\theta_{enc}$ by gradient descent according to Eqn. (5).
\State Update $\theta_{MAP}$ by gradient descent according to Eqn. (10).
\Until{Converge}
\State \textbf{Output}: $\theta_{enc}$, $\theta_{MAP}$.
\end{algorithmic}
\end{algorithm}

During pre-training stage, we minimize the combined loss: $L(\theta_{MAP}) + \lambda L(\theta_{enc})$. We use $\lambda$ to control the learning of MAP-Net.

\section{Experiments}
In this section, we empirically evaluate our proposed MAP-Net to show how much it can be used to improve the training efficiency of BERT. All codes are implemented based on fairseq in PyTorch toolkit\footnote{\url{https://github.com/pytorch/fairseq}}. For BERT,

\subsection{Experimental Design\label{sec:exprimentdesign}}

\begin{table*}[t]
    \centering 
    \caption{Hyperparameter for pretraining.} \vskip 0.1in
\begin{tabular}{lc}
\toprule
\textbf{Batch size} & 256 \\ \hline
\textbf{Sequence length} & 512 \\ \hline
\textbf{Training steps} & 1,000,000 \\ \hline
\textbf{Peak learning rate} & 1e-4 \\ \hline
\textbf{Adam eps} & 1e-6 \\ \hline
\textbf{Adam betas} & (0.9, 0.98) \\ \hline
\textbf{Warm-up steps} & 10,000 \\ \hline
\textbf{Learning rate decay} & Linear \\ \hline
\textbf{Dropout} & 0.1 \\ \hline
\textbf{Weight decay} & 0.01 \\ \bottomrule

\end{tabular}
    \label{tab:pt_space}
\end{table*}

We will first introduce the model architecture, and then introduce the detailed setups regarding the pre-training and the fine-tuning stage, respectively.

For the encoder, we set it as the same architecture as BERT \texttt{base} configuration (110M parameters), which is a 12-layer Transformer using GELU \cite{hendrycks2016gaussian} activation function. For each layer, the hidden size $H$ is set as 768 and the number of attention head $A$ is set as 12. For the Mask Proposal Network, we set its size to 1/2 of the encoder to improve efficiency. Specifically, we set the MAP-Net using a 12-layer Transformer with $H = 384$ and $A = 6$. We set $\lambda=1e \minus 2$ to trade-off the learning of the MAP-Net and that of the encoder.
Both the MAP-Net and the encoder use token embeddings as the input and we share these parameters between the two models.

\subsubsection{Pre-training}
\paragraph{Dataset} 
Following \cite{devlin2018bert}, we use English Wikipedia corpus\footnote{\url{https://dumps.wikimedia.org/enwiki}} and BookCorpus\footnote{As the dataset BookCorpus \cite{moviebook} is no longer freely distributed, we follow the suggestions from \cite{devlin2018bert} to crawl from \url{smashwords.com} and collect BookCorpus by ourselves. } for pre-training. By concatenating these two datasets, we obtain a corpus with roughly 3400M words in total. We follow a couple of consecutive pre-processing steps: segmenting documents into sentences by Spacy \footnote{\url{https://spacy.io}}, normalizing, lower-casing, and tokenizing the texts by Moses decoder \cite{Koehn2007MosesOS}, and finally, applying \textit{byte pair encoding} (BPE) \cite{DBLP:journals/corr/SennrichHB15} with setting the vocabulary size $|V|$ as 32,678.

\paragraph{Exploration v.s. Exploitation}
Ideally, the mask proposal network will interplay with the encoder from the beginning of the training. However, we observe that if the encoder purely uses the output of the mask proposal network from initialization, it cannot train well. We hypothesize this is because the mask network provides hard examples to the encoder, if the mask network is used from the beginning, the model always receive difficult masks which makes the optimization biased. To avoid such a problem and make the training efficient, we adopt an exploration-exploitation strategy: in each iteration, the masked sentences fed into the encoder are sampled from the uniform distribution with probability $p$ or MAP-Net with probability  $1-p$. The probability $p$ decreases linearly from 100\% at the beginning of training to 33\% at 1000k steps.

\paragraph{Optimization}
Following the standard settings used in many previous works \cite{devlin2018bert,liu2019roberta,clark2019electra}, we train the models for 1000$k$ steps with setting the batch size as 256 and the maximum sequence length as 512. For all the models to compare, we set the masked probability $p$ to be 0.15. For all experiments, we follow \cite{liu2019roberta} to replace 80\% of the masked positions by \texttt{[MASK]}, 10\% by randomly sampled words, and keep the remaining positions unchanged. But We choose the most widely used Adam \cite{DBLP:journals/corr/KingmaB14} as the optimizer, and set the hyper-parameter $\beta$ as $(0.9,0.98)$. The learning rate is set as 1e-4 with a 10$k$-step warm-up stage and then decays linearly to zero. We set the dropout probability as 0.1 and weight decay to 0.01. All models are run on 8 NVIDIA Tesla V100 GPUs. 

\subsubsection{Fine-tuning}

\begin{table*}[t]
    \centering 
    \caption{Hyperparameter search spaces for fine-tuning. Other hyperparameters are the same as in pretraining.} \vskip 0.1in
\begin{tabular}{lc}
\toprule
\textbf{Batch size} & \{16, 32\} \\ \hline
\textbf{Maximum epoch} & 10 \\ \hline
\textbf{Learning rate} & \{1e-5,~...,~8e-5\}  \\ \hline
\textbf{Warm-up ratio} & 0.06 \\ \hline
\textbf{Weight decay} & 0.1 \\ \bottomrule

\end{tabular}
    \label{tab:ds_space}
\end{table*}

\begin{table*}[ht]
    \centering
    \caption{The average score on GLUE tasks.} \vskip 0.1in
    \begin{tabular}{llcccccc}
    \toprule
        \multirow{2}*{\textbf{Task}} & \multirow{2}*{\textbf{Model}} & \multicolumn{4}{c}{\textbf{Pre-train steps}}  \\ \cline{3-6}
        ~ & ~ & 400k & 600k &  800k &  1000k \\ \hline
        \multirow{1}*{GLUE} & BERT & 80.86 & 81.41 & 81.62 & 81.68   \\
        \multirow{1}*{(all 9 tasks)} & MAP-Net & \textbf{81.15} & \textbf{81.48} & \textbf{82.11} & \textbf{81.88}  \\
    \bottomrule
    \end{tabular}
    \label{tab:main_res} 
\end{table*}

After the pretraining stage finishes, we discard the MAP-Net and only fine-tune the encoder on several downstream tasks. We use the GLUE (\textbf{G}eneral \textbf{L}anguage \textbf{U}nderstanding \textbf{E}valuation) dataset \cite{DBLP:journals/corr/abs-1804-07461} as the downstream tasks to evaluate the performance of the pre-trained models. Particularly, nine tasks within the GLUE dataset have been widely used for evaluation, including CoLA, RTE, MRPC, STS-B, SST-2, QNLI, QQP, and MNLI-m/MNLI-mm.

Same to the pre-training, we use Adam as the optimizer and set the hyper-parameter $\beta$ as $(0.9,0.98)$. Following all previous works, we apply the hyper-parameter search during the fine-tuning for each downstream task. The search space is listed in Table~\ref{tab:ds_space}. Each configuration will be run for ten times with different random seeds, and the average of these ten results on the development set will be used as the performance of one configuration. We will ultimately report the best number over all configurations.

\subsection{Experiment Results}

To fairly study the efficiency, besides of the last checkpoint (1000k steps), we also save some intermediate checkpoints. Specifically, we save additional checkpoints for 400k, 600k and 800k steps. And all the saved checkpoints will be used in various downstream fine-tune. And the results are shown in Table ~\ref{tab:main_res}.

\begin{figure*}
\centering
\includegraphics[scale=0.58]{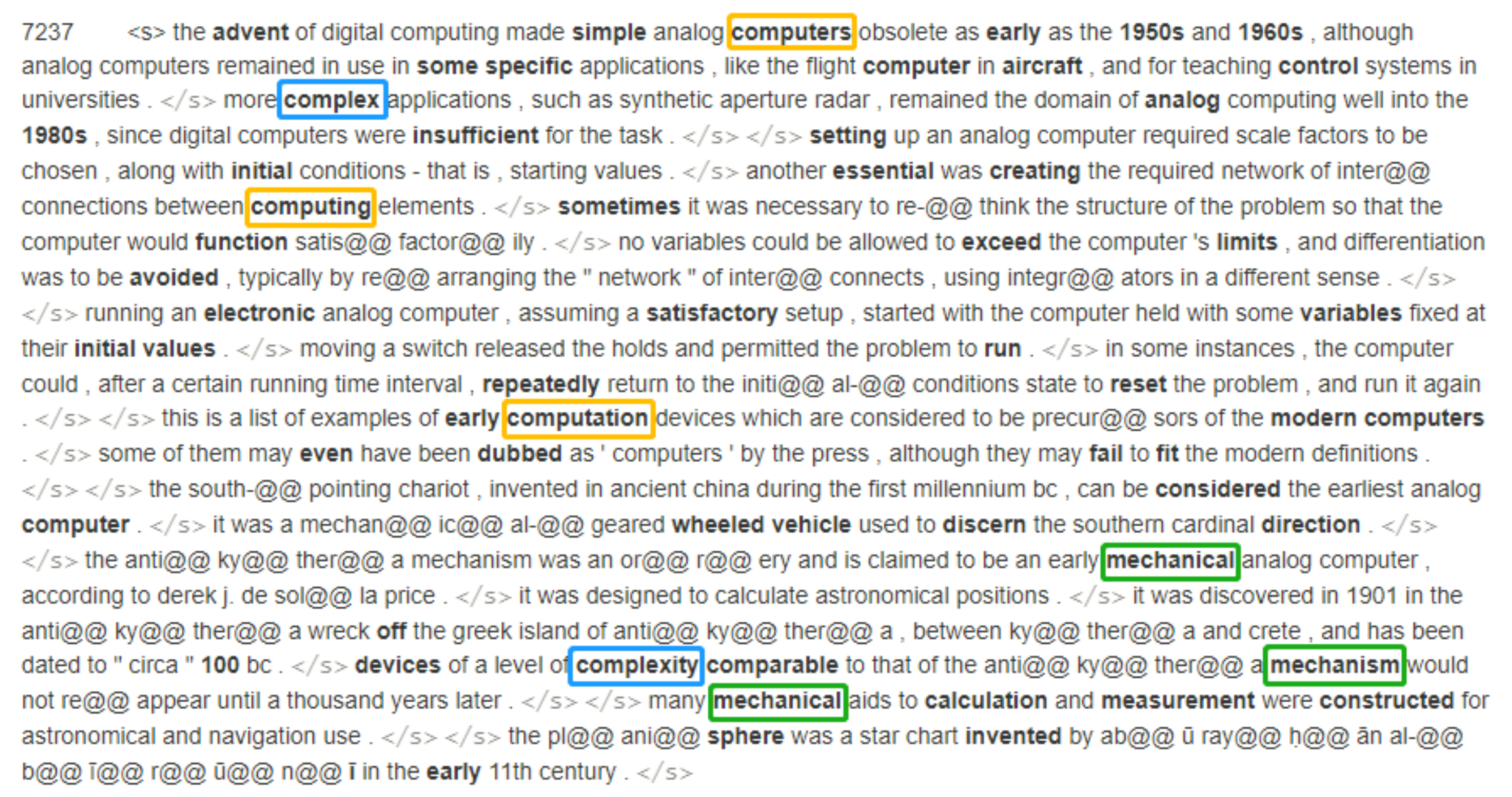}
\caption{A example sampled from the 600k checkpoint. The masked tokens are marked in bold and the patterns learned by MAP-Net are highlighted by small boxes of different colors.}     \label{fig:case_study}
\end{figure*}

\begin{figure*}
\centering
\includegraphics[scale=0.27]{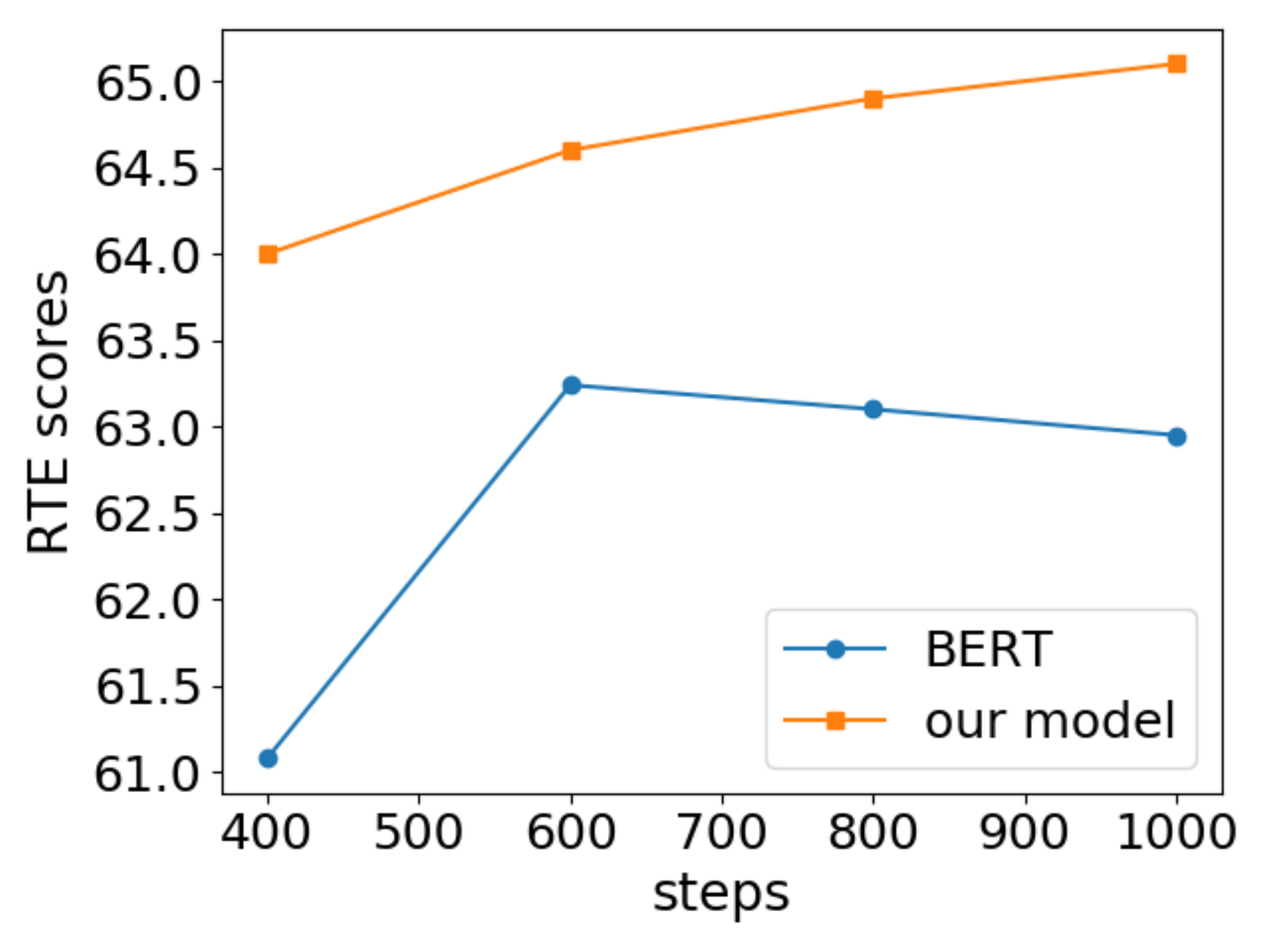}
\includegraphics[scale=0.27]{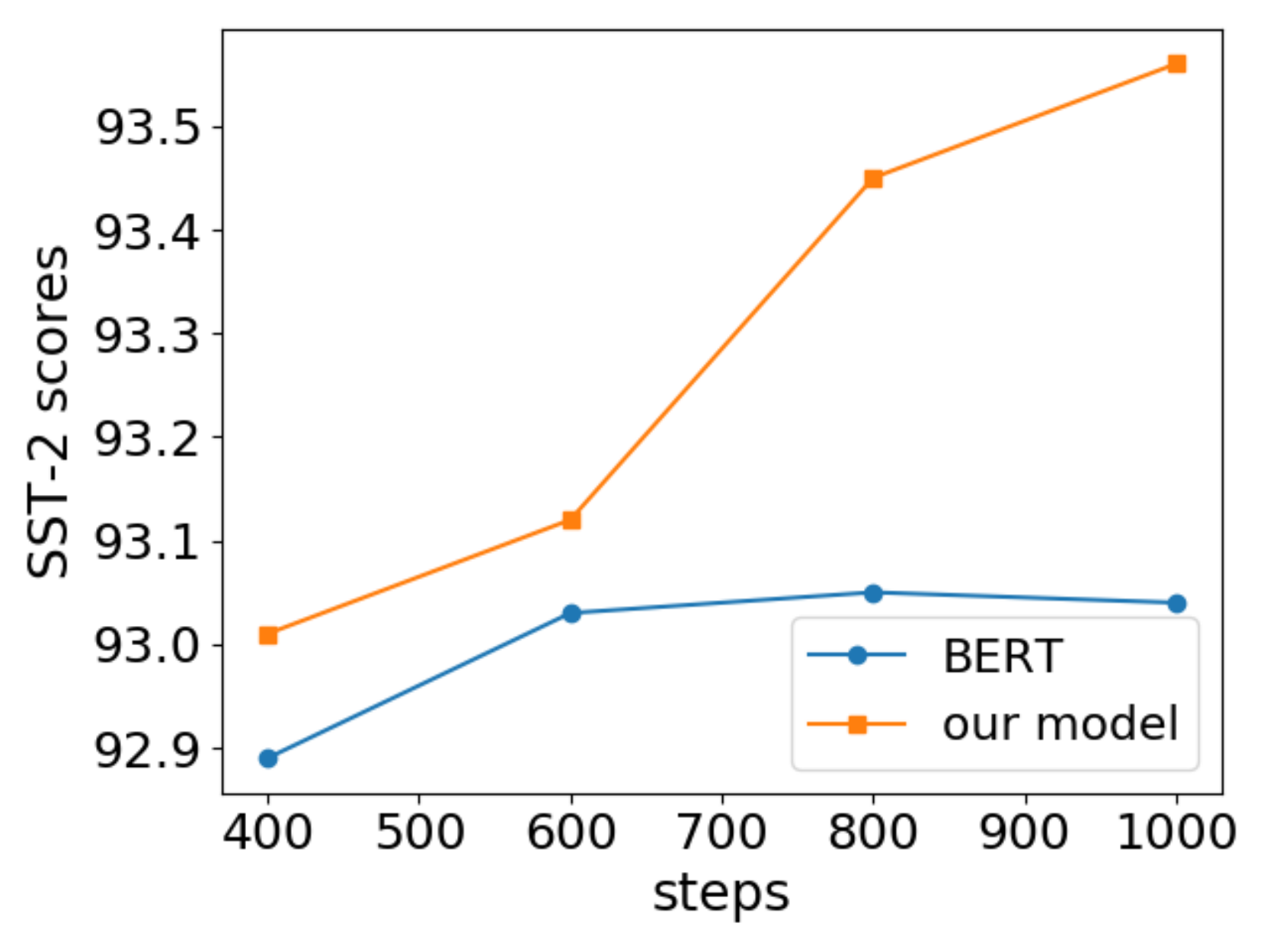}
\includegraphics[scale=0.27]{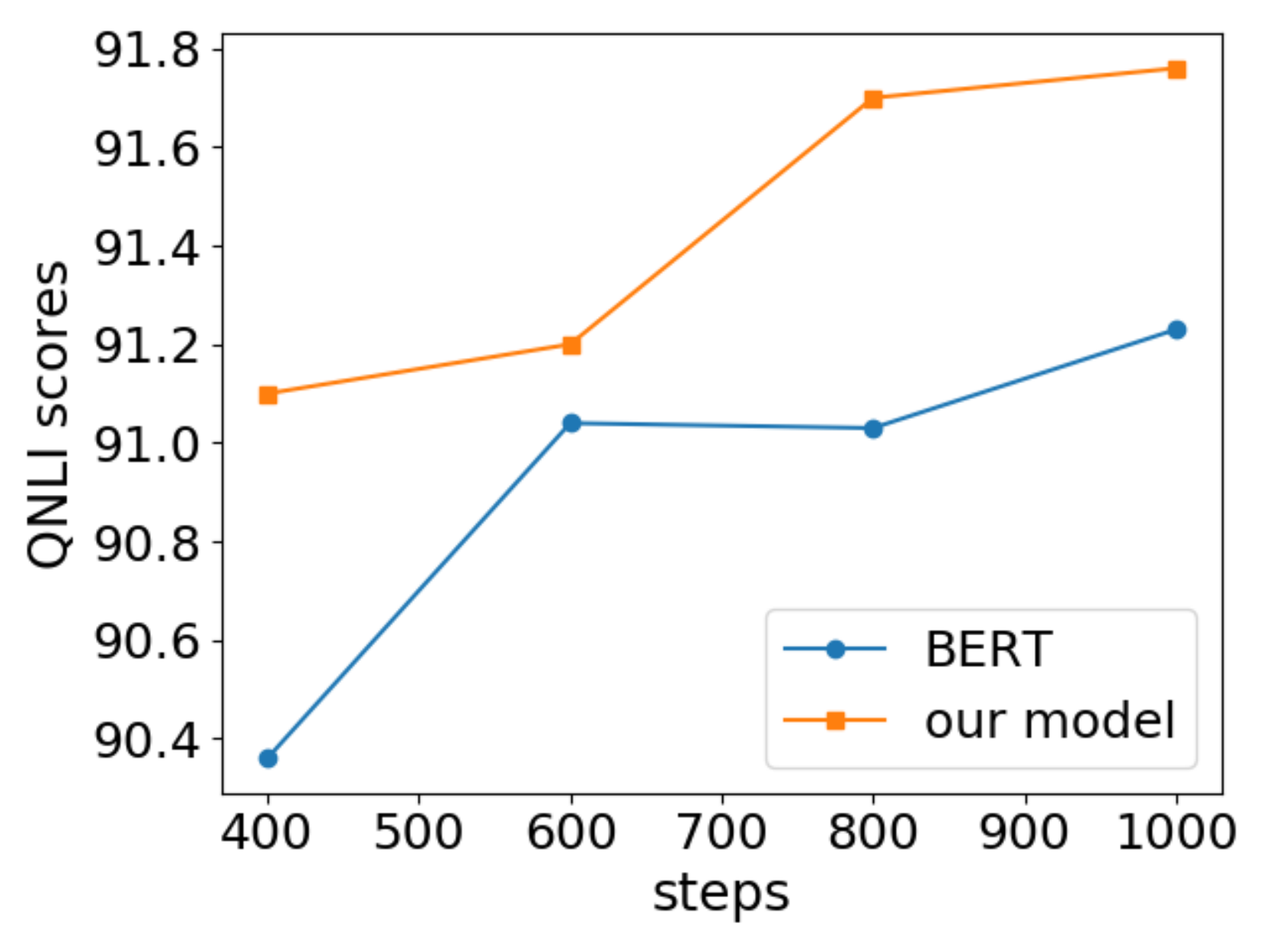}
\caption{The left, middle and right figures are the RTE, SST-2 and QNLI scores of different checkpoints respectively. The x-axis is the number of training steps, and the y-axis is the score of the task.}     \label{fig:tasks}
\end{figure*}

\paragraph{GLUE tasks} We report the average score of 9 GLUE tasks. As shown in Table~\ref{tab:main_res}, the proposed method can consistently outperform BERT in all checkpoints. For the best scores, the proposed method outperforms the baseline BERT by about 0.4 points. Besides, the result of the 600k checkpoint of the proposed method is almost pairing with the 1000k of baseline. These promising results demonstrate that the proposed MAP-Net can indeed help to speed up the pretraining, thanks to the gradient variance reduction. Due to space limitation, we only show the learning curves of three tasks, RTE, SST-2 and QNLI, in Figure \ref{fig:tasks} Left, Middle and Right. 
It is clear to see that for all tasks, our model converges much faster and achieves higher performance than the baseline BERT model. All these results demonstrate that the proposed MAP-Net yields great advantages in terms of efficiency and effectiveness compared to the baseline.

\paragraph{Case study} Finally, to further understand MAP-Net, we visualize the masked sentence sampled from the MAP-Net and the masked words are marked in bold. Due to space limitation, we only show one of the examples in Figure \ref{fig:case_study}. In spite of this, we can
see that the model tends to choose words such as nouns and verbs because they are more informative components of a sentence. More interestingly, in a sentence, our model can prevent information leakage by masking out different forms of the same word at the same time, such as "computer", "computing" and "computation", etc. So if BERT tries to restore them at the same time, it must learn better and deeper contextual information. This may be one of the reasons why our model performs better than BERT.

\section{Conclusion and Future Work}
In this work, we propose MAP-Net, which uses a mask proposal network to reduce the gradient variance of pre-training. In particular, given a sentence, the MAP-Net outputs a probability distribution over positions to sample masks. Then, the BERT model is trained to recover the masked sentence sampled from the MAP-Net, instead of uniform distribution. Extensive experiments demonstrate MAP-Net can do better on downstream tasks. It outperforms BERT on GLUE tasks with less training steps. In the future, we will continue exploring more methods to reduce the variance in pretraining, e.g., how to smartly select batched sentences.

\section*{Broader Impact}
Our method reduces the huge cost of pre-training, which can save energy and reduce greenhouse gas emissions, thereby contributing to global environmental protection.

\bibliographystyle{plain}  
\bibliography{references}  

\begin{thebibliography}{10}

\bibitem{alain2015variance}
Guillaume Alain, Alex Lamb, Chinnadhurai Sankar, Aaron Courville, and Yoshua
  Bengio.
\newblock Variance reduction in sgd by distributed importance sampling.
\newblock {\em arXiv preprint arXiv:1511.06481}, 2015.

\bibitem{clark2019electra}
Kevin Clark, Minh-Thang Luong, Quoc~V Le, and Christopher~D Manning.
\newblock Electra: Pre-training text encoders as discriminators rather than
  generators.
\newblock In {\em International Conference on Learning Representations}, 2019.

\bibitem{devlin2018bert}
Jacob Devlin, Ming-Wei Chang, Kenton Lee, and Kristina Toutanova.
\newblock Bert: Pre-training of deep bidirectional transformers for language
  understanding.
\newblock {\em arXiv preprint arXiv:1810.04805}, 2018.

\bibitem{dong2019unified}
Li~Dong, Nan Yang, Wenhui Wang, Furu Wei, Xiaodong Liu, Yu~Wang, Jianfeng Gao,
  Ming Zhou, and Hsiao-Wuen Hon.
\newblock Unified language model pre-training for natural language
  understanding and generation.
\newblock In {\em Advances in Neural Information Processing Systems}, pages
  13042--13054, 2019.

\bibitem{durrett2019probability}
Rick Durrett.
\newblock {\em Probability: theory and examples}, volume~49.
\newblock Cambridge university press, 2019.

\bibitem{gong2019efficient}
Linyuan Gong, Di~He, Zhuohan Li, Tao Qin, Liwei Wang, and Tieyan Liu.
\newblock Efficient training of bert by progressively stacking.
\newblock In {\em International Conference on Machine Learning}, pages
  2337--2346, 2019.

\bibitem{hendrycks2016gaussian}
Dan Hendrycks and Kevin Gimpel.
\newblock Gaussian error linear units (gelus).
\newblock {\em arXiv preprint arXiv:1606.08415}, 2016.

\bibitem{hochreiter1997long}
Sepp Hochreiter and J{\"u}rgen Schmidhuber.
\newblock Long short-term memory.
\newblock {\em Neural computation}, 9(8):1735--1780, 1997.

\bibitem{kalchbrenner2014convolutional}
Nal Kalchbrenner, Edward Grefenstette, and Phil Blunsom.
\newblock A convolutional neural network for modelling sentences.
\newblock {\em arXiv preprint arXiv:1404.2188}, 2014.

\bibitem{DBLP:journals/corr/KingmaB14}
Diederik~P. Kingma and Jimmy Ba.
\newblock Adam: {A} method for stochastic optimization.
\newblock {\em CoRR}, abs/1412.6980, 2014.

\bibitem{Koehn2007MosesOS}
Philipp Koehn, Hieu Hoang, Alexandra Birch, Chris Callison-Burch, Marcello
  Federico, Nicola Bertoldi, Brooke Cowan, Wade Shen, Christine Moran, Richard
  Zens, Chris Dyer, Ondrej Bojar, Alexandra Constantin, and Evan Herbst.
\newblock Moses: Open source toolkit for statistical machine translation.
\newblock In {\em ACL}, 2007.

\bibitem{liu2019roberta}
Yinhan Liu, Myle Ott, Naman Goyal, Jingfei Du, Mandar Joshi, Danqi Chen, Omer
  Levy, Mike Lewis, Luke Zettlemoyer, and Veselin Stoyanov.
\newblock Roberta: A robustly optimized bert pretraining approach.
\newblock {\em arXiv preprint arXiv:1907.11692}, 2019.

\bibitem{mikolov2013distributed}
Tomas Mikolov, Ilya Sutskever, Kai Chen, Greg~S Corrado, and Jeff Dean.
\newblock Distributed representations of words and phrases and their
  compositionality.
\newblock In {\em Advances in neural information processing systems}, pages
  3111--3119, 2013.

\bibitem{pennington2014glove}
Jeffrey Pennington, Richard Socher, and Christopher Manning.
\newblock Glove: Global vectors for word representation.
\newblock In {\em Proceedings of the 2014 conference on empirical methods in
  natural language processing (EMNLP)}, pages 1532--1543, 2014.

\bibitem{peters2018deep}
Matthew~E Peters, Mark Neumann, Mohit Iyyer, Matt Gardner, Christopher Clark,
  Kenton Lee, and Luke Zettlemoyer.
\newblock Deep contextualized word representations.
\newblock {\em arXiv preprint arXiv:1802.05365}, 2018.

\bibitem{radford2018improving}
Alec Radford, Karthik Narasimhan, Tim Salimans, and Ilya Sutskever.
\newblock Improving language understanding by generative pre-training.
\newblock {\em URL https://s3-us-west-2. amazonaws.
  com/openai-assets/research-covers/language-unsupervised/language\_
  understanding\_paper. pdf}, 2018.

\bibitem{radford2019language}
Alec Radford, Jeffrey Wu, Rewon Child, David Luan, Dario Amodei, and Ilya
  Sutskever.
\newblock Language models are unsupervised multitask learners.
\newblock {\em OpenAI Blog}, 1(8), 2019.

\bibitem{DBLP:journals/corr/SennrichHB15}
Rico Sennrich, Barry Haddow, and Alexandra Birch.
\newblock Neural machine translation of rare words with subword units.
\newblock {\em CoRR}, abs/1508.07909, 2015.

\bibitem{socher2011dynamic}
Richard Socher, Eric~H Huang, Jeffrey Pennin, Christopher~D Manning, and
  Andrew~Y Ng.
\newblock Dynamic pooling and unfolding recursive autoencoders for paraphrase
  detection.
\newblock In {\em Advances in neural information processing systems}, pages
  801--809, 2011.

\bibitem{strubell2019energy}
Emma Strubell, Ananya Ganesh, and Andrew McCallum.
\newblock Energy and policy considerations for deep learning in nlp.
\newblock {\em arXiv preprint arXiv:1906.02243}, 2019.

\bibitem{tai2015improved}
Kai~Sheng Tai, Richard Socher, and Christopher~D Manning.
\newblock Improved semantic representations from tree-structured long
  short-term memory networks.
\newblock {\em arXiv preprint arXiv:1503.00075}, 2015.

\bibitem{vaswani2017attention}
Ashish Vaswani, Noam Shazeer, Niki Parmar, Jakob Uszkoreit, Llion Jones,
  Aidan~N Gomez, {\L}ukasz Kaiser, and Illia Polosukhin.
\newblock Attention is all you need.
\newblock In {\em Advances in Neural Information Processing Systems}, pages
  5998--6008, 2017.

\bibitem{DBLP:journals/corr/abs-1804-07461}
Alex Wang, Amanpreet Singh, Julian Michael, Felix Hill, Omer Levy, and
  Samuel~R. Bowman.
\newblock {GLUE:} {A} multi-task benchmark and analysis platform for natural
  language understanding.
\newblock {\em CoRR}, abs/1804.07461, 2018.

\bibitem{yang2019xlnet}
Zhilin Yang, Zihang Dai, Yiming Yang, Jaime Carbonell, Ruslan Salakhutdinov,
  and Quoc~V Le.
\newblock Xlnet: Generalized autoregressive pretraining for language
  understanding.
\newblock {\em arXiv preprint arXiv:1906.08237}, 2019.

\bibitem{you2019large}
Yang You, Jing Li, Sashank Reddi, Jonathan Hseu, Sanjiv Kumar, Srinadh
  Bhojanapalli, Xiaodan Song, James Demmel, and Cho-Jui Hsieh.
\newblock Large batch optimization for deep learning: Training bert in 76
  minutes.
\newblock {\em arXiv preprint arXiv:1904.00962}, 1(5), 2019.

\bibitem{moviebook}
Yukun Zhu, Ryan Kiros, Richard Zemel, Ruslan Salakhutdinov, Raquel Urtasun,
  Antonio Torralba, and Sanja Fidler.
\newblock Aligning books and movies: Towards story-like visual explanations by
  watching movies and reading books.
\newblock In {\em arXiv preprint arXiv:1506.06724}, 2015.

\end{thebibliography}


\end{document}